\documentclass{article}

\usepackage{microtype}
\usepackage{graphicx}
\usepackage{xcolor}
\usepackage{subfigure}
\usepackage{booktabs} 
\usepackage[inline]{enumitem}
\usepackage{multirow}
\usepackage{algorithm}
\usepackage{algorithmic}

\usepackage{mathtools}

\usepackage{Definitions}
\usepackage{hyperref}
\hypersetup{colorlinks, citecolor={blue}}

\DeclarePairedDelimiter\ceil{\lceil}{\rceil}
\DeclarePairedDelimiter\floor{\lfloor}{\rfloor}

\usepackage[accepted]{icml2020}

\newcommand{\modelshort}{BiGG}
\newcommand{\modelfull}{BIg Graph Generation}

\icmltitlerunning{Scalable Deep Generative Modeling for Sparse Graphs}

\begin{document}

\twocolumn[
\icmltitle{Scalable Deep Generative Modeling for Sparse Graphs}




\begin{icmlauthorlist}
\icmlauthor{Hanjun Dai}{brain}
\icmlauthor{Azade Nazi}{brain}
\icmlauthor{Yujia Li}{dm}
\icmlauthor{Bo Dai}{brain}
\icmlauthor{Dale Schuurmans}{brain}
\end{icmlauthorlist}

\icmlaffiliation{brain}{Google Research, Brain Team}
\icmlaffiliation{dm}{DeepMind}

\icmlcorrespondingauthor{Hanjun Dai}{hadai@google.com}

\icmlkeywords{Machine Learning, ICML}

\vskip 0.3in
]



\printAffiliationsAndNotice{}  

\setlength{\abovedisplayskip}{3pt}
\setlength{\abovedisplayshortskip}{3pt}
\setlength{\belowdisplayskip}{3pt}
\setlength{\belowdisplayshortskip}{3pt}
\setlength{\jot}{2pt}
\setlength{\floatsep}{1ex}
\setlength{\textfloatsep}{1ex}
\setlength{\intextsep}{1ex}
\setlength{\parskip}{1ex}

\begin{abstract}

Learning graph generative models is a challenging task for deep learning and has wide applicability to a range of domains like chemistry, biology and social science. However current deep neural methods suffer from limited scalability: for a graph with $n$ nodes and $m$ edges, existing deep neural methods require $\Omega(n^2)$ complexity by building up the  adjacency matrix. On the other hand, many real world graphs are actually sparse in the sense that $m\ll n^2$. Based on this, we develop a novel autoregressive model, named BiGG, that utilizes this sparsity to avoid generating the full adjacency matrix, and importantly reduces the graph generation time complexity to $O((n + m)\log n)$. Furthermore, during training this autoregressive model can be parallelized with $O(\log n)$ synchronization stages, which makes it much more efficient than other autoregressive models that require $\Omega(n)$. Experiments on several benchmarks show that the proposed approach not only scales to orders of magnitude larger graphs than previously possible with deep autoregressive graph generative models, but also yields better graph generation quality.

\end{abstract}

\section{Introduction}
\label{sec:intro}

Representing a distribution over graphs
provides a principled foundation for tackling many important problems
in knowledge graph completion~\citep{xiao2016transg},
de novo drug design~\citep{li2018learning, simonovsky2018graphvae},
architecture search~\citep{xie2019exploring}
and program synthesis~\citep{brockschmidt2018generative}.
The effectiveness of graph generative modeling
usually depends on \emph{learning}
the distribution given a collection of relevant training graphs.
However, training a generative model over graphs
is usually quite difficult due to their discrete and combinatorial nature.

Classical generative models of graphs,
based on random graph theory
\cite{erdHos1960evolution,barabasi1999emergence,watts1998collective},
have been long studied
but typically only capture a small set of specific graph properties,
such as degree distribution.
Despite their computational efficiency, 
these distribution models are usually too inexpressive 
to yield competitive results in challenging applications.

Recently, deep graph generative models that exploit the increased
capacity of neural networks to learn more expressive graph distributions
have been successfully applied to real-world tasks.
Prominent examples include
VAE-based methods
\cite{kipf2016variational,simonovsky2018graphvae},
GAN-based methods \cite{bojchevski2018netgan},
flow models \cite{liu2019graph,shi2020graphaf}
and autoregressive models
\cite{li2018learning,you2018graphrnn,liao2019efficient}.
Despite the success of these approaches in modeling small graphs,
e.g.\ molecules with hundreds of nodes,
they are not able to scale to graphs with over 10,000 nodes.

A key shortcoming of current deep graph generative models is that they
attempt to generate a \textit{full} graph adjacency matrix,
implying a computational cost of $\Omega(n^2)$ for
a graph with $n$ nodes and $m$ edges.
For large graphs, it is impractical to sustain such a
quadratic time and space complexity,
which creates an inherent trade-off between expressiveness and efficiency.
To balance this trade-off, 
most recent work has introduced various conditional independence assumptions
\cite{liao2019efficient},
ranging from the fully auto-regressive but slow GraphRNN \cite{you2018graphrnn}, to the fast but fully factorial GraphVAE \cite{simonovsky2018graphvae}.

In this paper, we propose an alternative approach that does not commit to 
explicit conditional independence assumptions,
but instead exploits the fact that most interesting
real-world graphs are \emph{sparse},
in the sense that $m\ll n^2$.
By leveraging sparsity,
we develop a new graph generative model, \modelshort{} (\modelfull{}),
that streamlines the generative process and avoids explicit consideration
of every entry in an adjacency matrix.
The approach is based on three key elements:
(1) an $O(\log n)$ process for generating each edge using a binary tree
data structure, inspired by R-MAT \cite{chakrabarti2004r};
(2) a tree-structured autoregressive model for generating the set of edges
associated with each node; and
(3) an autoregressive model defined over the sequence of nodes.
By combining these elements,
\modelshort{} can generate a sparse graph in $O((n + m)\log n)$ time,
which is a substantial improvement over $\Omega(n^2)$.

For training,
the design of \modelshort{} 
also
allows every context embedding in the
autoregressive model to be computed  
in only
$O(\log n)$ sequential steps,
which enables 
significant gains in training speed through parallelization.
By comparison,
the context embedding in GraphRNN requires $O(n^2)$ sequential steps to
compute, while GRAN requires $O(n)$ steps.
In addition, we develop a training mechanism that only requires sublinear
memory cost,
which in principle makes it possible to train models of graphs with
millions of nodes on a single GPU. 

On several benchmark datasets,
including synthetic graphs and real-world graphs of proteins,
3D mesh and SAT instances,
\modelshort{} is able to achieve comparable or superior sample quality
than the previous state-of-the-art,
while being orders of magnitude more scalable. 

To summarize the main contributions of this paper:
\begin{itemize}[leftmargin=*,nolistsep,nosep]
	\item We propose an autoregressive generative model,
	\modelshort{}, that 
can generate sparse graphs
in $O((n+m)\log n)$ time,
successfully modeling
graphs with 100k nodes on 1 GPU.
	\item The training process can be largely parallelized,
	requiring only $O(\log n)$ steps to synchronize learning updates.
	\item Memory cost is reduced to $O(\sqrt{m \log n})$
	for training and $O(\log n)$ for inference. 
	\item \modelshort{} not only scales to orders of magnitude larger graphs
	than current deep models, it also yields comparable or better model quality
	on several benchmark datasets. 
\end{itemize}

\noindent\textbf{Other related work}
There has been a lot of work \citep{chakrabarti2004r, robins2007exprandom, Leskovec2010kronecker, Airoldi2008mixed} on generating graphs with a set of specific properties like degree distribution, diameter, and eigenvalues. All these classical models are hand-engineered to model a particular family of graphs, and thus not general enough.
Besides the general graphs, a lot of work also exploit domain knowledge for better performance in specific domains.  Examples of this include \citep{kusner2017grammar,dai2018syntax,jin2018junction,liu2018constrained} for modeling molecule graphs, and \cite{you2019g2sat} for SAT instances.
See~\appref{app:related} for more discussions.

\section{Model}
\label{sec:model}

A graph $G = (V, E)$ is defined by a set of nodes $V$ and set of edges
$E\subseteq V\times V$,
where a tuple $e_i =(u, v) \in E$ is used to represent an edge between
node $u$ and $v$.
We denote $n=|V|$ and $m=|E|$ as the number of nodes and edges in
$G$ respectively.
Note that a given graph may have multiple equivalent adjacency matrix
representations,
with different node orderings.
However, given an ordering of the nodes $\pi$, there is a one-to-one mapping
between the graph structure $G$ and the adjacency matrix
$A^\pi\in\{0,1\}^{n\times n}$.

Our goal is to learn a generative model, $p(G)$, given a set of training
graphs
$\Dcal_{train} = \cbr{G_1, G_2, \ldots, G_{|\Dcal_{train}|}}$.
In this paper, we assume the graphs are not attributed,
and focus on the graph topology only.
Such an assumption implies 
\begin{align}
	p\rbr{G} &= p(V)p(E|V) = p(|V|=n) \sum_\pi p(E,\pi|n) \nonumber \\
		&= p(|V|=n)\sum_\pi p(A^\pi),
\end{align}
where $p(E,\pi|n)$ is the probability of generating the set of edges $E$
under a particular ordering $\pi$,
which is
equivalent to the probability of a particular adjacency matrix $A^\pi$.  
Here,
$p(|V|=n)$ is the distribution of number of nodes in a graph.
In this paper, we use a single canonical ordering $\pi(G)$ to model each
graph $G$, as in \cite{li2018learning}:
\begin{equation}
    p(G)\simeq p(|V|=n) p(A^{\pi(G)}),
\end{equation}
which is clearly a lower bound on $p(G)$~\citep{liao2019efficient}.   
We estimate $p(|V|=n)$ directly using the empirical distribution
over graph size, and only learn an expressive model for $p(A)$.
In the following presentation, we therefore omit the ordering $\pi$
and assume a default canonical ordering of nodes in the graph when appropriate.

As $A$ will be extremely sparse for large sparse graphs ($m\ll  n^2$),
generating only the non-zero entries in $A$, \ie, the edge set $E$, 
will be much more efficient than the full matrix:
\begin{equation}
	p(A) = p(e_1) p(e_2|e_1) \ldots p(e_m|e_1, \ldots, e_{m-1}),
\end{equation}
where each $e_i=(u,v)$ include the indices of the two nodes associated with one edge, resulting in a generation process of $m$ steps.
We order the edges following the node ordering, hence this process generates
all the edges for the first row in $A$, before generating the second row,
etc.
A naive approach to generating a single edge 
will be $O(n)$ if we factorize $p(e_i) = p(u) p(v|u)$
and assume both $p(u)$ and $p(v|u)$ to be simple multinomials over $n$ nodes.
This, however, will not give us any benefit over traditional methods.


\subsection{Recursive edge generation}
\label{sec:edge_gen}

The main scalability bottleneck is the large output space.
One way to reduce the output space size is to use a hierarchical
recursive decomposition, inspired by the classic random graph model R-MAT
\cite{chakrabarti2004r}.
In R-MAT, each edge is put into the adjacency matrix by dividing the 2D
matrix into 4 equally sized quadrants,
and recursively descending into one of the quadrants,
until reaching a single entry of the matrix.
In this generation process, the complexity of generating each edge is only $O(\log n)$.

\begin{figure}[t]
\centering
\includegraphics[width=0.48\textwidth]{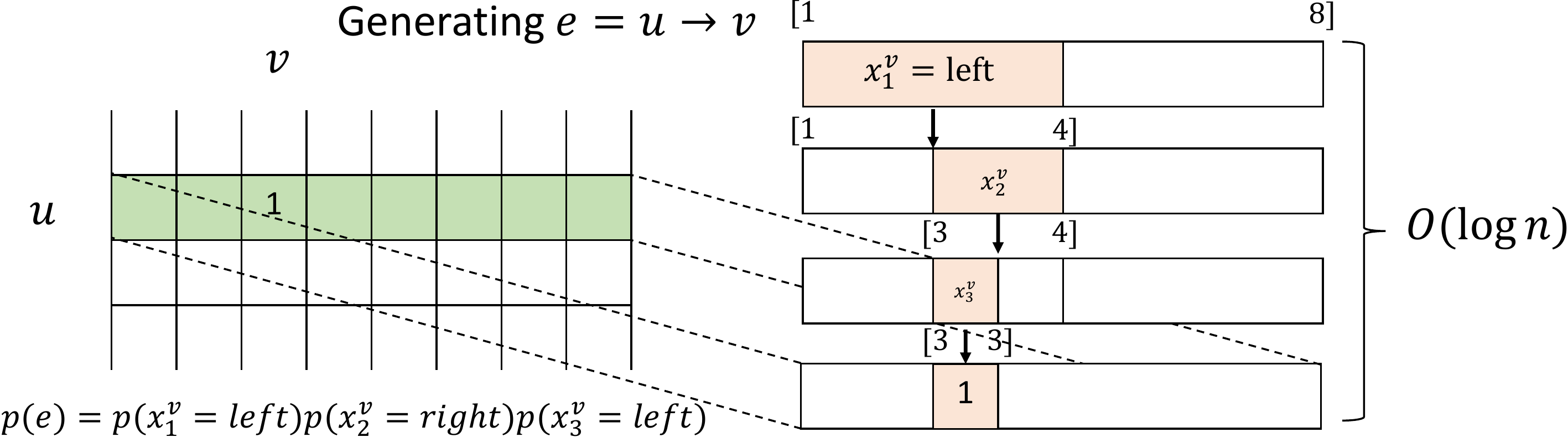}
\caption{Recursive generation of the edge $e=(u, v)$ given $u$. \label{fig:edge_gen}}
\end{figure}

We adopt the recursive decomposition design of R-MAT, and further simplify and neuralize it, to make our model efficient and expressive.  In our model, we generate edges following the node ordering row by row,
so that each edge only needs to be put into the right place in a single row,
reducing the process to 1D, as illustrated in \figref{fig:edge_gen}.
For any edge $(u,v)$, the process of picking node $v$ as one of $u$'s neighbors
starts by dividing the node index interval $[1, n]$ in half,
then recursively descending into one half until reaching a single entry.
Each $v$ corresponds to a unique sequence of decisions $x^v_1, ..., x^v_d$, where $x^v_i\in\{\text{left}, \text{right}\}$ is the $i$-th decision in the sequence, and $d=\ceil{\log_2 n}$ is the maximum number of required decisions to specify $v$.

The probability of $p(v|u)$ can then be formulated as
\begin{equation}
\textstyle
    p(v|u) = \prod_{i=1}^{\ceil{\log_2 n}} p(x_i = x^v_i),
\end{equation}
where each $p(x_i=x^v_i)$ is the probability of following the decision that leads to $v$ at step $i$.

Let us use $E_u = \cbr{(u, v)\in E}$ to denote the set of edges incident to node $u$, and $\Ncal_u = \{v|(u,v)\in E_u\}$. Generating only a single edge is similar to hierarchical softmax~\citep{mnih2009scalable}, and applying the above procedure repeatedly can generate all of $|\Ncal_u|$ edges in $O(|\Ncal_u|\log n)$ time. But we can do better than that when generating all these edges. 

\textbf{Further improvement using binary trees. }
As illustrated in the left half of \figref{fig:row_gen},
the process of jointly generating all of $E_u$ is equivalent to building up a binary tree $\Tcal_u$, where each tree node $t\in\Tcal_u$ corresponds to a graph node index interval $[v_l, v_r]$, and for each $v\in\Ncal_u$ the process starts from the root $[1, n]$ and ends in a leaf $[v, v]$.

Taking this perspective, we propose a more efficient generation process for
$E_u$, which generates the tree directly instead of repeatedly generating
each leaf through a path from the root.
We propose a recursive process that builds up the tree following a depth-first
or in-order traversal order, where we start at the root, and recursively for
each tree node $t$:
(1) decide if $t$ has a left child denoted as $\lch(t)$, and
(2) if so recurse into $\lch(t)$ and generate the left sub-tree, and then
(3) decide if $t$ has a right child denoted as $\rch(t)$,
(4) if so recurse into $\rch(t)$ and generate the right sub-tree, and
(5) return to $t$'s parent.
This process is shown in \algref{alg:row_gen},
which will be elaborated in next section.

Overloading the notation a bit, we use $p(\lch(t))$ to denote the
probability that tree node $t$ has a left child,
when $\lch(t)\ne \emptyset$, or does not have a left child,
when $\lch(t)=\emptyset$, under our model,
and similarly define $p(\rch(t))$.
Then the probability of generating $E_u$ or equivalently tree $\Tcal_u$ is
\begin{equation}
    p(E_u) = p(\Tcal_u) = \prod_{t\in\Tcal_u} p(\lch(t)) p(\rch(t))
.
\end{equation}
This new process generates each tree node exactly once,
hence the time complexity is proportional to the tree size $O(|\Tcal_u|)$,
and it is clear that $|\Tcal_u| \le |\Ncal_u|\log n$,
since $|\Ncal_u|$ is the number of leaf nodes in the tree
and $\log n$ is the max depth of the tree.
The time saving comes from removing the duplicated effort near the root of the tree.  When $|\Ncal_u|$ is large, \ie{} as some fraction of $n$ when $u$ is one of the ``hub'' nodes in the graph, the tree $\Tcal_u$ becomes dense and our new generation process will be significantly faster, as the time complexity becomes close to $O(n)$ while generating each leaf from the root would require
$\Omega(n\log n)$ time.

In the following, we present our approach to make this model fully
autoregressive, \ie\ making $p(\lch(t))$ and $p(\rch(t))$ depend on all the decisions made so far in the process of generating the graph, and make this model neuralized so that all the probability values in the model come from expressive deep neural networks.

\subsection{Autoregressive conditioning for generating $\Tcal_u$}
\label{sec:row_gen}

\begin{algorithm}[t]
\begin{algorithmic}[1]
\FUNCTION{recursive($u, t, h_u^{top}(t)$)}{
	\IF{is\_leaf(t)}{
		\STATE Return $\vec{1}, \cbr{\text{edge index that $t$ represents}}$
	}
	\ENDIF
	\STATE has\_left $\sim p(\text{lch}_u(t) | h_u^{top}(t))$ using Eq.~\eqref{eq:plch}
	\IF{has\_left}
		\STATE Create $lch_u(t)$
		, and let $h_u^{bot}(\text{lch}_u(t)), \Ncal_u^{l, t} \leftarrow $ {\bf recursive}($u, \text{lch}_u(t), h_u^{top}(lch_u(t))$)
	\ELSE
		\STATE $h_u^{bot}(\text{lch}_u(t)) \leftarrow \vec{0}, \Ncal_u^{l, t} = \emptyset$
	\ENDIF
	\STATE has\_right $\sim p(\text{rch}_u(t) | \hat{h}_u^{top}(rch_u(t)))$ using Eq.~\eqref{eq:prch}
	\IF{has\_right}
		\STATE Create $rch_u(t)$, and let $h_u^{bot}(\text{rch}_u(t)), \Ncal_u^{r, t} \leftarrow $ {\bf recursive}($u, \text{rch}_u(t), h_u^{top}(rch_u(t)))$)
	\ELSE
		\STATE $h_u^{bot}(\text{ rch}_u(t)) \leftarrow \vec{0}, \Ncal_u^{r, t} = \emptyset$
	\ENDIF
	\STATE 	$h_u^{bot}(t) = \text{TreeCell}^{bot}(h_u^{bot}(\text{lch}_u(t)), h_u^{bot}(\text{rch}_u(t)))$
	\STATE $\Ncal_u^t = \Ncal_u^{l, t} \cup \Ncal_u^{r, t}$
	\STATE Return $h_u^{bot}(t)$, $\Ncal_u^t$
}
\ENDFUNCTION
\end{algorithmic}
    \caption{Generating outgoing edges of node $u$}
    \label{alg:row_gen}
\end{algorithm}

In this section we consider how to add autoregressive conditioning to
$p(\lch(t))$ and $p(\rch(t))$ when generating $\Tcal_u$.

\begin{figure}[t]
\centering
\includegraphics[width=0.48\textwidth]{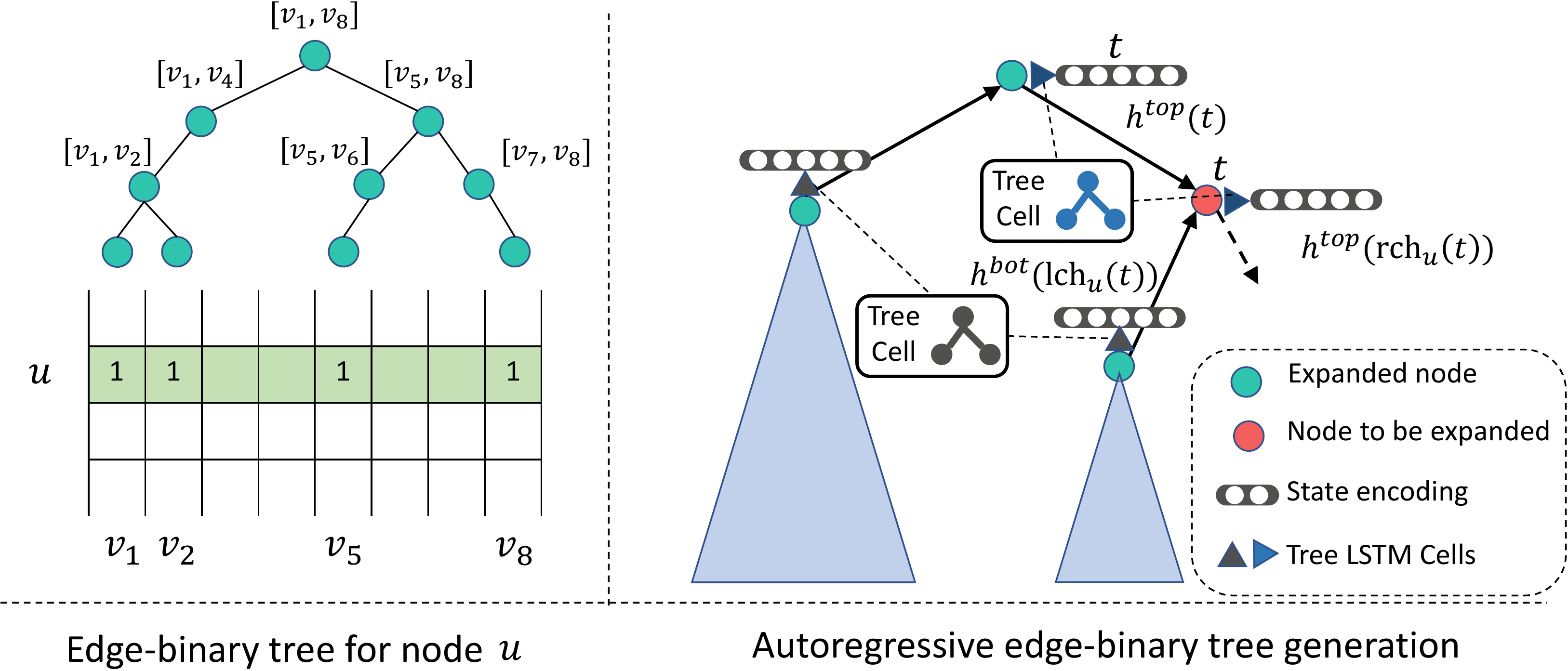}
\caption{Autoregressive generation of edge-binary tree of node $u$. To generate the red node $t$, the two embeddings that captures $t$'s left subtree (orange region) and nodes with in-order traversal before $t$ (blue region) respectively are used for conditioning. \label{fig:row_gen}}
\end{figure}

In our generation process, the decision about whether $\lch(t)$ exists for a particular tree node $t$ is made after $t$, all its ancestors, and all the left sub-trees of the ancestors are generated.  We can use a top-down context vector $h^{top}_u(t)$ to summarize all these contexts, and modify $p(\lch(t))$ to $p(\lch(t)|h^{top}_u(t))$.  Similarly, the decision about $\rch(t)$ is made after generating $\lch(t)$ and its dependencies,
and $t$'s entire left-subtree (see \figref{fig:row_gen} right half for illustration).  We therefore need both the top-down context $h^{top}_u(t)$, as well as the bottom-up context $h^{bot}_u(\lch(t))$ that summarizes the sub-tree rooted at $\lch(t)$, if any.  The autoregressive model for $p(\Tcal_u)$ therefore becomes
\begin{align}
    p(\Tcal_u) = \prod_{t\in\Tcal_u} & p(\lch(t)|h^{top}_u(t)) \nonumber\\
    & p(\rch(t)|h^{top}_u(t), h^{bot}_u(\lch(t))),
\end{align}
and we can recursively define
\begin{align}
    h^{bot}_u(t) =& \TreeCell^{bot}(h^{bot}_u(\lch(t)), h^{bot}_u(\rch(t))) \nonumber \\
    h^{top}_u(\lch(t)) =& \LSTMCell(h^{top}_u(t), \embed(\text{left})) \label{eq:tree_embed} \\
    \hat{h}^{top}_u(\rch(t)) =& \TreeCell^{top}(h^{bot}_u(\lch(t)), h^{top}_u(\lch(t))) \nonumber \\
    h^{top}_u(\rch(t)) =& \LSTMCell(\hat{h}^{top}_u(\rch(t)), \embed(\text{right})), \nonumber
\end{align}
where $\TreeCell^{bot}$ and $\TreeCell^{top}$ are two TreeLSTM cells~\citep{tai2015improved} that combine information
from the incoming nodes into a single node state, and $\embed(\text{left})$ and $\embed(\text{right})$ represents the embedding vector for the binary values ``left'' and ``right''.  We initialize $h^{bot}_u(\emptyset) = \vec{0}$, and discuss $h^{top}_u(\text{root})$ in the next section.

The distributions can then be parameterized as
\begin{align}
    p(\lch(t)|\cdot) &= \text{Bernoulli}(\sigma(W_l^\top h^{top}_u(t) + b_l)) \label{eq:plch}
,
\\
    p(\rch(t)|\cdot) &= \text{Bernoulli}(\sigma(W_r^\top \hat{h}^{top}_u(\rch(t)) + b_r))
.
\label{eq:prch}
\end{align}

\subsection{Full autoregressive model}
\label{sec:full_gen}

With the efficient recursive edge generation and autoregressive conditioning presented in \secref{sec:edge_gen} and \secref{sec:row_gen} respectively, we are ready to present the full autoregressive model for generating the entire adjacency matrix $A$. 

The full model will utilize the autoregressive model for $\Ncal_u$ as building blocks. Specifically, we are going to generate the adjacency matrix $A$ row by row: 
\begin{eqnarray}
	p(A) = p(\cbr{\Ncal_u}_{u \in V}) = \prod_{u\in V} p\rbr{\Ncal_u| \cbr{\Ncal_{u'}: u' < u}}
.
\end{eqnarray}
Let $g_u^0 = h_u^{bot}(t_1)$ be the embedding that summarizes $\Tcal_u$, suppose we have an efficient mechanism to encode $\sbr{g_1, g_2, \ldots, g_u}$ into $h^{row}_u$, then we can effectively use $h^{row}_{u-1}$ to generate $\Tcal_u$ and thus the entire process would become autoregressive. Again, since there are $n$ rows in total, using a chain structured LSTM would make the history length too long for large graphs.
Therefore, we use an approach
inspired by the Fenwick tree~\citep{fenwick1994new} which is a data structure that maintains the prefix sum efficiently. Given an array of numbers, the Fenwick tree allows calculating any prefix sum or updating any single entry in $O(\log L)$ for a sequence of length $L$. We build such a tree  to maintain and update the \textit{prefix embeddings}. We denote it as row-binary forest as such data structure is a forest of binary trees. 

\begin{figure}[t]
\centering
\includegraphics[width=0.48\textwidth]{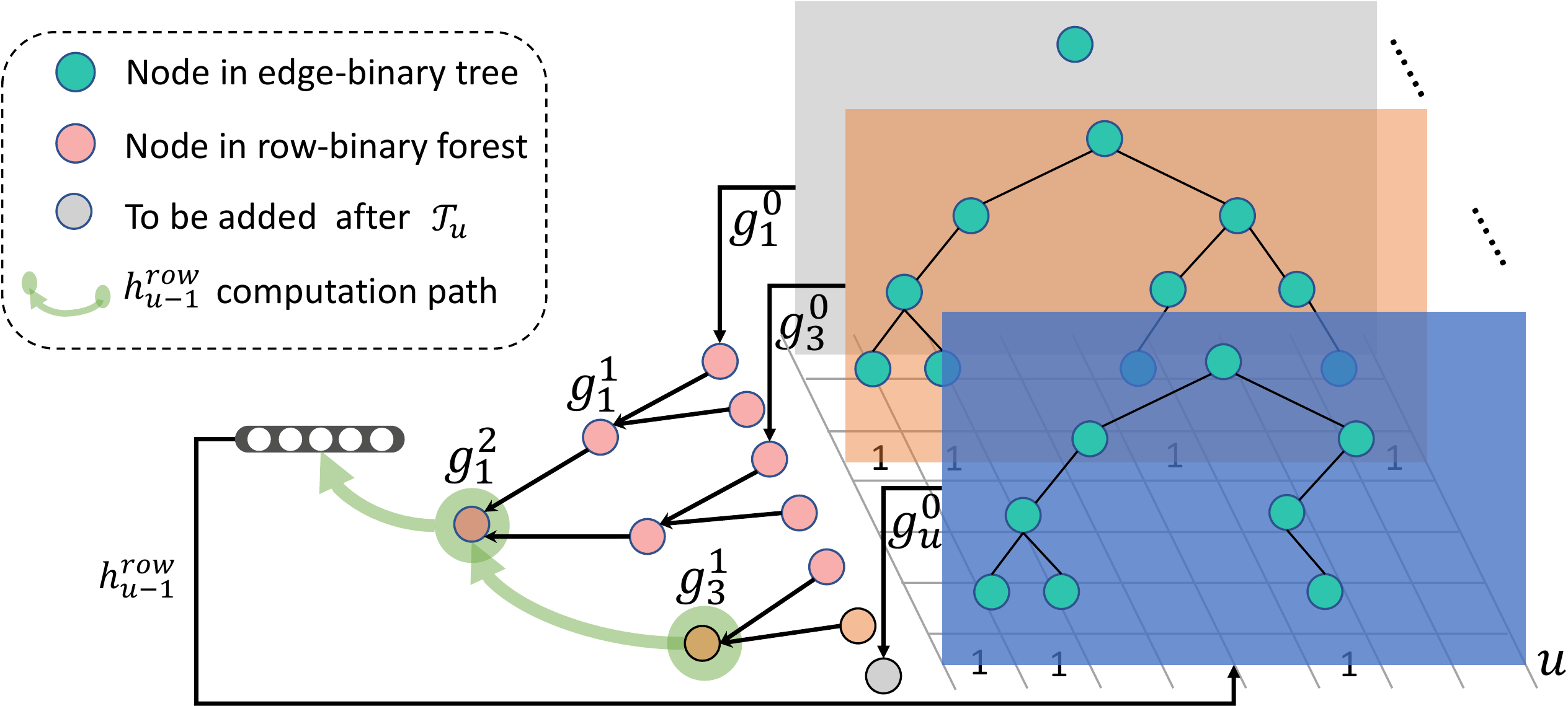}
\caption{Autoregressive conditioning across rows of adjacency matrix. Embedding $h^{row}_{u-1}$ summarizes all the rows before $u$, and is used for generating $\Tcal_u$ next. \label{fig:row2048} }
\end{figure}

\figref{fig:row2048} demonstrates one solution. Before generating the edge-binary tree $\Tcal_u$, the embeddings that summarize each individual edge-binary tree $\Rcal_u = \cbr{\Tcal_{u'}: u' < u}$ will be organized into the row-binary forest $\Gcal_u$. This forest is organized into $\floor*{\log (u-1)} + 1$ levels, with the bottom $0$-th level as edge-binary tree embeddings. Let $g^i_j \in \Gcal_u$ be the $j$-th node in the $i$-th level, then
\begin{equation}
\label{eq:row_forest}
	g_j^i = \text{TreeCell}^{row}(g_{j*2-1}^{i-1}, g_{j*2}^{i-1}),
\end{equation}
where $0 \leq i \leq \floor*{\log (u-1)} + 1, 1 \leq j \leq \floor*{\frac{|\Rcal_u|}{2^i}}$.

\noindent\textbf{Embedding row-binary forest} One way to embed this row-binary forest is to embed the root of each tree in this forest. As there will be at most one root in each level (otherwise the two trees in the same level will be merged into a larger tree for the next level), the number of tree roots will also be $O(\log n)$ at most. Thus we calculate $h^{row}_{u}$ as follows:
\begin{equation}
	h^{row}_{u} = \text{LSTM}\rbr{\sbr{g^i_{\floor*{\frac{u}{2^i}}}, \text{ where } u \text{ $\&$ } 2^i = 2^i}}
	\label{eq:h_row}
.
\end{equation}
Here, $\&$ is the bit-level `and' operator. Intuitively as in Fenwick tree, the calculation of any prefix sum of length $L$ requires the block sums that corresponds to each binary digit in the binary bits representation of integer $L$. 
 Recall the operator $h^{top}_u(\cdot)$ defined in \secref{sec:row_gen}, here $h^{top}_u(root) = h^{row}_{u-1}$ when $u > 1$, and equals to zero when $u = 1$. With the embedding of $\Gcal_u$ at each state served as `glue', we can connect all the individual row generation modules in an autoregressive way. 

\noindent\textbf{Updating row-binary forest} It is also efficient to update such forest every time a new $g_u^0$ is obtained after the generation of $\Tcal_u$. Such an updating procedure is similar to the Fenwick tree update. 
As each level of this forest has at most one root node, merging in the way defined in \eqref{eq:row_forest} will happen at most once per each level,
which effectively makes the updating cost to be $O(\log n)$. 

\begin{algorithm}[t]
\begin{algorithmic}[1]
\FUNCTION{update\_forest($u, \Gcal_{u-1}, g_u^0$)}{
	\STATE $\Gcal_{u} = \Gcal_{u-1} \cup \cbr{g_u^0}$
	\FOR{$i \leftarrow 0$ to $\floor*{\log(u - 1)}$}{
		\STATE $j \leftarrow \arg\max_j \II\sbr{g^i_j \in \Gcal_{u}}$
		\IF{such $j$ exists and $j$ is an even number}{
		\STATE $g^{i+1}_{j / 2} \leftarrow \text{TreeCell}^{row}(g^i_{j-1}, g^i_j)$
		\STATE $\Gcal_{u} \leftarrow \Gcal_{u} \cup g^{i+1}_{j / 2}$
		}
		\ENDIF
	}
	\ENDFOR
	\STATE Update $h_u^{row}$ using Eq~\eqref{eq:h_row}
	\STATE Return: $h_u^{row}, \Gcal_{u}$, $\Rcal_{u-1} \cup \cbr{g_u^0}$
}
\ENDFUNCTION
\item[]
\STATE Input: The number of nodes $n$
\STATE $h_0^{row} \leftarrow \vec{0}, \Rcal_0 = \emptyset, \Gcal_0 = \emptyset, E = \emptyset$
\FOR{$u \leftarrow 1$ to $n$}{
	\STATE Let $t_1$ be the root of an empty edge-binary tree
	\STATE $g_u^0, \Ncal_u \leftarrow$ {\bf recursive}$(u, t_1, h_{u-1}^{row})$
	\STATE $h_u^{row}, \Gcal_u, \Rcal_u \leftarrow$ {\bf update\_forest}($u, \Rcal_{u-1}, \Gcal_{u-1}, g_u^0$)
}
\ENDFOR
\STATE Return: $G$ with $V=\cbr{1, \ldots, n}$ and $E = \cup_{u=1}^n \Ncal_u$
\end{algorithmic}
    \caption{Generating graph using \modelshort{}}
    \label{alg:graph_gen}
\end{algorithm}

\algref{alg:graph_gen} summarizes the entire procedure for sampling
a graph from our model in a fully autoregressive manner.

\vspace{-2mm}
\begin{theorem}
	\modelshort{} generates a graph with $n$ node and $m$ edges in $O\rbr{(n+m) \log n}$ time. In the extreme case where $m \simeq n^2$, the overall complexity becomes $O(n^2)$. 
\end{theorem}
\vspace{-2mm}
\begin{proof}
The generation of each edge-binary tree in~\secref{sec:row_gen} requires time complexity proportional to the number of nodes in the tree.
The Fenwick tree query and update both take $O(\log n)$ time,
hence maintaining the data structure takes $O(n\log n)$.
The overall complexity is $O\rbr{n\log n + \sum_{u=1}^n |\Tcal_u|}$.
For a sparse graph $|\Tcal_u| = O(|\Ncal_u| \log n)$,
hence $\sum_{u=1}^n |\Tcal_u| = \sum_{u=1}^n |\Ncal_u| \log n = O(m\log n)$.
For a complete graph, where $m = O(n^2)$, each $\Tcal_u$ will be a full
binary tree with $n$ leaves, hence $|\Tcal_u| = 2n-1$ and the
overall complexity would be $O(n\log n + n^2) = O(n^2)$. 
\end{proof}
\vspace{-6mm}

\section{Optimization}
\label{sec:opt}

In this section, we propose several ways to scale up the training of our auto-regressive model. For simplicity, we focus on how to speed up the training with a single graph. Training multiple graphs can be easily extended.

\subsection{Training with $O(\log n)$ synchronizations}
\label{sec:par_train}

A classical autoregressive model like LSTM is fully sequential,
which allows no concurrency across steps.
Thus training LSTM with a sequence of length $L$ takes $\Omega(L)$
of synchronized computation.
In this section, we show how to exploit the characteristic of
\modelshort{} to increase concurrency during training.  

\begin{figure}[t]
\centering
\includegraphics[width=0.48\textwidth]{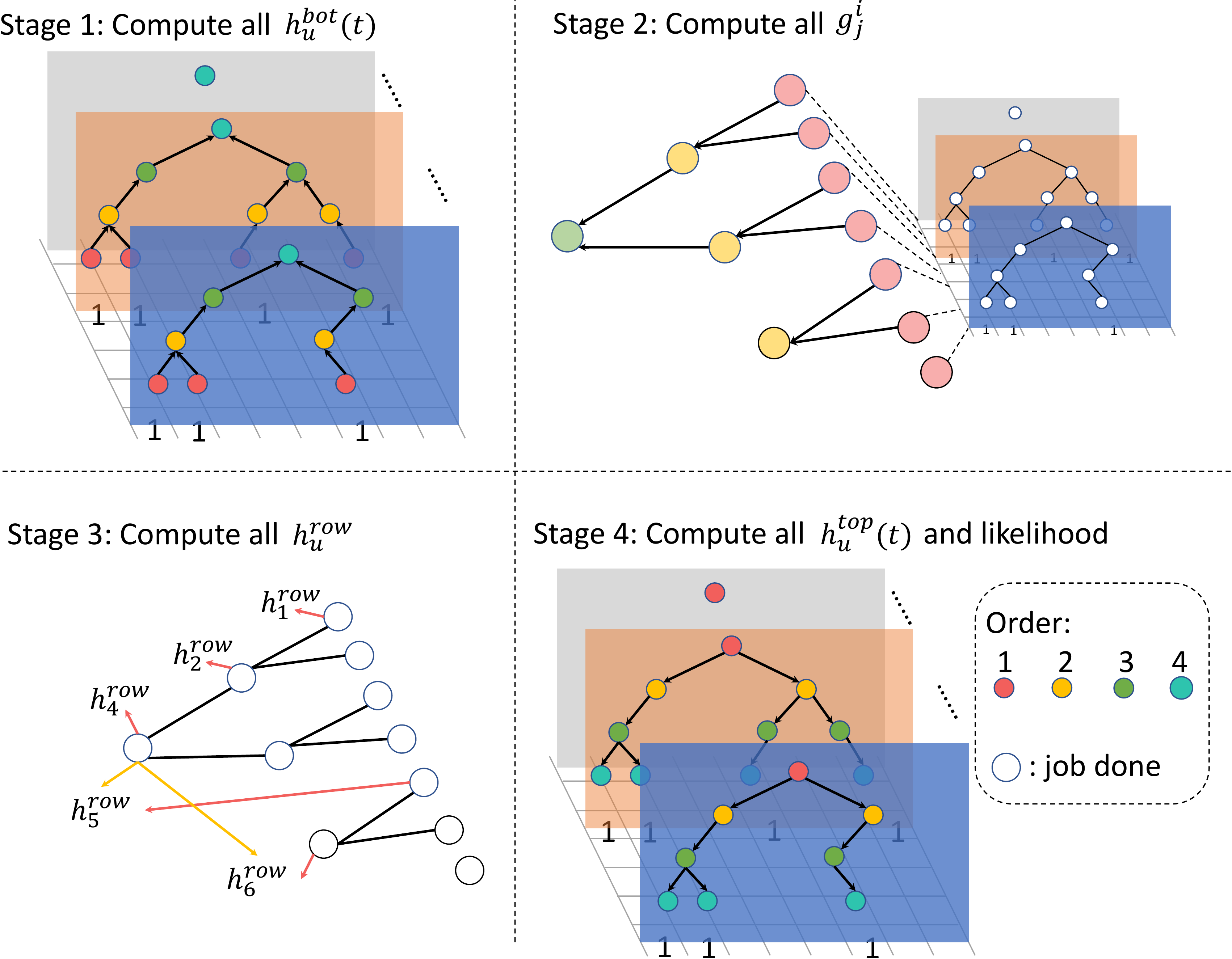}
\caption{Parallelizing computation during training. The four stages are executed sequentially. In each stage, the embeddings of nodes with the same color can be computed concurrently. \label{fig:par_train}}
\end{figure}

Given a graph $G$, estimating the likelihood under the current model can be divided into four steps. 
\begin{enumerate}[wide,noitemsep,topsep=0.5pt,parsep=0pt,partopsep=0.5pt]
	\item \textbf{Computing $h_u^{bot}(t), \forall u \in V, t \in \Tcal_u$ }:
as the graph is given during training, the corresponding edge-binary trees $\cbr{\Tcal_u}$ are also known.
From Eq~\eqref{eq:tree_embed} we can see that the embeddings $h_u^{bot}(t)$
of all nodes at the same depth of tree can be computed concurrently
without dependence, and the synchronization only happens between different
depths.
Thus $O(\log n)$ steps of synchronization is sufficient. 
	\item \textbf{Computing $g_j^i \in \Gcal_{n}$}: the row-binary forest grows monotonically, thus the forest $\Gcal_{n}$ in the end contains all the embeddings needed for computing $\cbr{h_u^{row}}$. Similarly, computing $\cbr{g_j^i}$ synchronizes $O(\log n)$ steps. 
	\item \textbf{Computing $h_u^{row}, \forall u \in V$}:
as each $h_u^{row}$ runs an LSTM independently,
this stage simply runs LSTM on a batch of $n$ sequences with length $O(\log n)$. 
	\item \textbf{Computing $h_u^{top}$ and likelihood}: the last step computes the likelihood using Eq~\eqref{eq:plch} and ~\eqref{eq:prch}.
This is similar to the first step, except that the computation happens
in a top-down direction in each $\Tcal_u$. 
\end{enumerate}

\figref{fig:par_train} demonstrates this process.
In summary, the four stages each take $O(\log n)$ steps of synchronization.
This allows us to train large graphs much more efficiently
than a simple sequential autoregressive model.

\subsection{Model parallelism}
\label{sec:model_par}

It is possible that during training the graph is too large to fit into memory.
Thus to train on large graphs, model parallelism is more important than
data parallelism.

To split the model, as well as intermediate computations,
into different machines,
we divide the adjacency matrix into multiple consecutive chunks of rows,
where each machine is responsible for one chunk.
It is easy to see that by doing so, Stage 1 and Stage 4 mentioned in
\secref{sec:par_train} can be executed concurrently on all the machines
without any synchronization,
as the edge-binary trees can be processed independently once the
conditioning states like $\cbr{h_u^{row}}$ are made ready by synchronization.

\begin{figure}[h]
\centering
\includegraphics[width=0.48\textwidth]{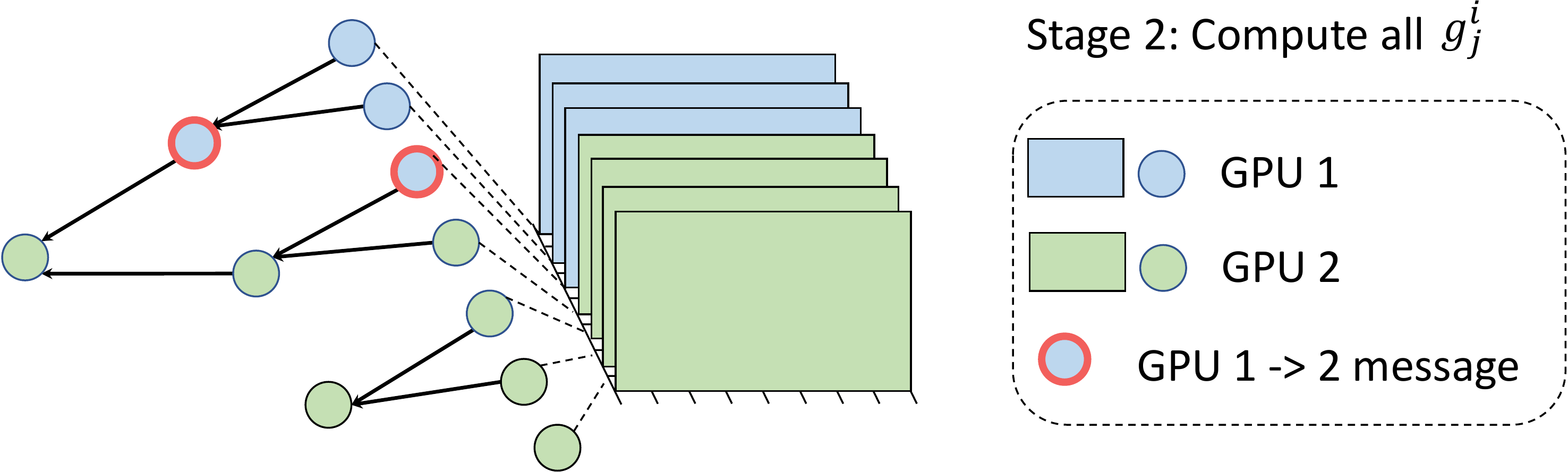}
\caption{Model parallelism for training single graph. Red circled nodes are computed on GPU 1 but is required by GPU 2 as well. \label{fig:model_par}}
\end{figure}
 
\figref{fig:model_par} illustrates the situation when training a graph with 7 nodes using 2 GPUs. During Stage 1, GPU 1 and 2 work concurrently to compute $\cbr{g_u^0}_{u=1}^3$ and $\cbr{g_u^0}_{u=4}^7$, respectively. In Stage 2, the embeddings $g_1^1$ and $g_3^0$ are needed by GPU 2 when computing $g_1^2$ and $g_2^1$. We denote such embeddings as `g-messages'.
Note that such `g-messages' will transit in the opposite direction
when doing a backward pass in the gradient calculation. 
Passing `g-messages' introduces serial dependency across GPUs.
However as the number of such embeddings is upper bounded by $O(\log n)$ depth
of row-binary forest,
the communication cost is manageable. 
\subsection{Reducing memory consumption}
\label{sec:sqrtn}

\noindent\textbf{Sublinear memory cost:} 
Another way to handle the memory issue when training large graphs
is to recompute certain portions of the hidden layers in the neural network
when performing backpropagation, to avoid storing such layers during 
the forward pass. 
\citet{chen2016training} introduces a computation scheduling mechanism
for sequential structured neural networks that achieves $O(\sqrt L)$
memory growth for an $L$-layer neural network. 

We divide rows as in~\secref{sec:model_par}.
During the forward pass, only the `g-messages' between chunks are kept.
The only difference from the previous section is that
the edge-binary tree embeddings are recomputed,
due to the single GPU limitation.
The memory cost will be:
\begin{equation}
	O(\max\cbr{k\log n, \frac{m}{k}})
\end{equation}
Here $O(k\log n)$ accounts for the memory holding the `g-message', and $O(\frac{m}{k})$ accounts for the memory of $\Tcal_u$ in each chunk.
The optimal $k$ is achieved when $k\log n = \frac{m}{k}$, hence
$k = O(\sqrt \frac{m}{\log n})$ and the corresponding memory cost is
$O(\sqrt{m \log n})$.
Also note that such sublinear cost requires only one additional feedforward 
in Stage 1, so this will not hurt much of the training speed.

\noindent\textbf{Bits compression:}
The vector $h_u^{bot}(t)$ summarizes the edge-binary tree structure rooted at node $t$ for $u$-th row in adjacency matrix $A$, as defined in Eq~\eqref{eq:tree_embed}. As node $t$ represents the interval $[v_l, v_r]$ of the row, another equivalent way is to directly use $A[u, v_l:v_r]$, \ie, the binary vector to represent $h_u^{bot}(t)$. Each $h_u^{bot}(t')$ where $t' = [v'_l, v'_r] \subset t = [v_l, v_r]$ is also a binary vector. Thus no neural network computation is needed in the subtree rooted at node $t$. Suppose we use such bits representation for any nodes that have the corresponding interval length no larger than $L$, then for a full edge-binary tree $\Tcal_u$ (\ie, $u$ connects to every other node in graph) which has $2n-1$ nodes in the tree, the corresponding storage required for neural part is $\ceil{2 \frac{n}{L} - 1}$ which essentially reduces the memory consumption of neural network to $\frac{1}{L}$ of the original cost. Empirically we use $L=256$ in all experiments, which saves $50\%$ of the memory during training without losing any information in representation.

Note that to represent an interval $A[u, v_l:v_r]$ of length $b = v_r - v_l + 1 \leq L$, we use vector $\vb \in \cbr{-1, 0, 1}^{L}$ where 
\begin{equation}
\nonumber
	\vb = \sbr{\underbrace{-1, \ldots, -1}_{L - b}, \underbrace{A[u, v_l], A[u, v_l+1], \ldots, A[u, v_r]}_{b}}
\end{equation}
That is to say, we use ternary bit vector to encode both the interval length and the binary adjacency information. 

\subsection{Position encoding:}
During generation of $\cbr{\Tcal_u}$, each tree node $t$ of the edge-binary tree knows the span $[v_l, v_r]$ which corresponds to the columns it will cover. One way is to augment $h^{top}_u(t)$ with the position encoding as:
\begin{equation}
	\hat{h}^{top}_u(t) = h^{top}_u(t) + \text{PE}(v_r - v_l)
\end{equation}
where PE is the position encoding using sine and cosine functions of different frequencies as in~\citet{vaswani2017attention}.
Similarly, the $h_u^{row}$ in Eq~\eqref{eq:h_row} can be augmented by $\text{PE}(n - u)$ in a similar way. With such augmentation, the model will know more context into the future, and thus help improve the generative quality. 

 Please refer to our released open source code located at \url{https://github.com/google-research/google-research/tree/master/bigg} for more implementation and experimental details.

\vspace{-2mm}
\section{Experiment}
\label{sec:experiment}
\vspace{-1mm}

\subsection{Model Quality Evaluation on Benchmark Datasets}
\label{sec:exp_benchmark}

In this part, we compare the quality of our model with previous work on a set of benchmark datasets.
We present results on median sized general graphs with number of nodes ranging in 0.1k to 5k in \secref{sec:exp_general_graph}, and on large SAT graphs with up to 20k nodes in \secref{sec:exp_sat_graph}. In \secref{sec:ablation} we perform ablation studies of~\modelshort{} with different sparsity and node orders.

\subsubsection{General graphs}
\label{sec:exp_general_graph}

\begin{table*}[t]
\centering
\resizebox{1.0\textwidth}{!}{%
\begin{tabular}{cccccccc}
	\toprule
	\multirow{2}{*}{Datasets} & & \multicolumn{6}{c}{Methods} \\
	\cmidrule(lr){3-8}
	& & Erdos-Renyi & GraphVAE & GraphRNN-S & GraphRNN & GRAN & \modelshort\ \\
	\midrule
	\multirow{2}{*}{Grid} & Deg. & 0.79 & 7.07$e^{-2}$ & 0.13 & 1.12$e^{-2}$ & 8.23$e^{-4}$ & $\mathbf{4.12e^{-4}}$ \\
	& Clus. & 2.00 & 7.33$e^{-2}$ & 3.73$e^{-2}$ & 7.73$e^{-5}$ & 3.79$e^{-3}$ & $\mathbf{7.25e^{-5}}$ \\
	$|V|_{max}=361$, $|V|_{avg} \approx 210$ & Orbit & 1.08 & 0.12 & 0.18 & 1.03$e^{-3}$ & 1.59$e^{-3}$ & $\mathbf{5.10e^{-4}}$ \\
	$|E|_{max}=684$, $|E|_{avg} \approx 392$ & Spec. & 0.68 & 1.44$e^{-2}$ & 0.19 & 1.18$e^{-2}$ & 1.62$e^{-2}$ & $\mathbf{9.28e^{-3}}$ \\
	\midrule
	\multirow{2}{*}{Protein} & Deg. & 5.64$e^{-2}$ & 0.48 & 4.02$e^{-2}$ & 1.06$e^{-2}$ & 1.98$e^{-3}$ & $\mathbf{9.51e^{-4}}$ \\
	& Clus. & 1.00 & 7.14$e^{-2}$ & 4.79$e^{-2}$ & 0.14 & 4.86$e^{-2}$ & $\mathbf{2.55e^{-2}}$ \\
	$|V|_{max}=500$, $|V|_{avg} \approx 1575$ & Orbit & 1.54 & 0.74 & 0.23 & 0.88 & 0.13 & $\mathbf{2.26e^{-2}}$ \\
	$|E|_{max}=258$, $|E|_{avg} \approx 646$ & Spec. & 9.13$e^{-2}$ & 0.11 & 0.21 & 1.88$e^{-2}$ & 5.13$e^{-3}$ & $\mathbf{4.51e^{-3}}$ \\
	\midrule
	\multirow{2}{*}{3D Point Cloud} & Deg. & 0.31 & OOM & OOM & OOM & 1.75$e^{-2}$ & $\mathbf{2.56e^{-3}}$ \\
	& Clus. & 1.22 & OOM & OOM & OOM & 0.51 & {\bf 0.21} \\
	$|V|_{max}=5037$, $|V|_{avg} \approx 1377$ & Orbit & 1.27 & OOM & OOM & OOM & 0.21 & $\mathbf{7.18e^{-3}}$ \\
	$|E|_{max}=10886$, $|E|_{avg} \approx 3074$ & Spec. & 4.26$e^{-2}$ & OOM & OOM & OOM & 7.45$e^{-3}$ & $\mathbf{3.40e^{-3}}$ \\
	\midrule
	\multirow{2}{*}{Lobster} & Deg. &  0.24 & 2.09$e^{-2}$ & 3.48$e^{-3}$ & 9.26$e^{-5}$ & 3.73$e^{-2}$ & $\mathbf{2.94e^{-5}}$ \\
	& Clus. & 3.82$e^{-2}$ & 7.97$e^{-2}$ & 4.30$e^{-2}$ & {\bf 0.00} & {\bf 0.00} & {\bf 0.00} \\
	$|V|_{max}=100$, $|V|_{avg} \approx 53$ & Orbit & 2.42$e^{-2}$ & 1.43$e^{-2}$ & 2.48$e^{-4}$ & 2.19$e^{-5}$ & 7.67$e^{-4}$ & $\mathbf{1.51e^{-5}}$ \\
	\multirow{2}{*}{$|E|_{max}=99$, $|E|_{avg} \approx 52$} & Spec. & 0.33 & 3.94$e^{-2}$ & 6.72$e^{-2}$ & 1.14$e^{-2}$ & 2.71$e^{-2}$ & $\mathbf{8.57e^{-3}}$ \\
	& Err. & 1.00 & 0.91 & 1.00 & {\bf 0.00} & 0.12 & {\bf 0.00} \\
	\bottomrule
\end{tabular}%
}
\vspace{-2mm}
\caption{Performance on benchmark datasets. The MMD metrics uses test functions from $\cbr{\text{Deg., Clus., Orbit., Spec.}}$.  For all the metrics, the smaller the better. Baseline results are obtained from~\citet{liao2019efficient}, where OOM indicates the out-of-memory issue. \label{tab:benchmark}}
\end{table*}

The general graph benchmark is obtained from~\citet{liao2019efficient} and part of it was also used in \cite{you2018graphrnn}. This benchmark has four different datasets: (1) Grid, 100 2D grid graphs; (2) Protein, 918 protein graphs~\citep{dobson2003distinguishing}; (3) Point cloud, 3D point clouds of 41 household objects~\citep{neumann2013graph}; (4) Lobster, 100 random Lobster graphs~\citep{golomb1996polyominoes}, which are trees where each node is at most 2 hops away from a backbone path. \tabref{tab:benchmark} contains some statistics about each of these datasets.
We use the same protocol as~\citet{liao2019efficient} that splits the graphs into training and test sets. 

\noindent\textbf{Baselines:} We compare with  deep generative models including GraphVAE~\citep{simonovsky2018graphvae},  GraphRNN, GraphRNN-S \citep{you2018graphrnn} and GRAN ~\citep{liao2019efficient}.
We also include the Erdős–Rényi random graph model that only estimates the edge density.
Since our setups are exactly the same, 
the baseline results are directly copied from~\citet{liao2019efficient}.

\noindent\textbf{Evaluation:} We use exactly the same evaluation metric as~\citet{liao2019efficient}, which compares the distance between the distribution of held-out test graphs and the generated graphs. We use maximum mean discrepancy (MMD) with four different test functions, namely the node degree, clustering coefficient, orbit count and the spectra of the graphs from the eigenvalues of the normalized graph Laplacian. Besides the four MMD metrics, we also use the error rate for Lobster dataset. This error rate reflects the fraction of generated graphs that doesn't have Lobster graph property. 

\noindent\textbf{Results:} \tabref{tab:benchmark} reports the results on all the four datasets. We can see the proposed \modelshort{} outperforms all other methods on all the metrics. The gain becomes more significant on the largest dataset, \ie, the 3D point cloud. While GraphVAE and GraphRNN gets out of memory, the orbit metric of~\modelshort{} is 2 magnitudes better than GRAN. This dataset reflects the scalability issue of existing deep generative models. Also from the Lobster graphs we can see, although GRAN scales better than GraphRNN, it yields worse quality due to its approximation of edge generation with mixture of conditional independent distributions. Our~\modelshort{} improves the scalability while also maintaining the expressiveness.

\subsubsection{SAT graphs}
\label{sec:exp_sat_graph}

\begin{table*}[t]
\centering
\resizebox{1.0\textwidth}{!}{%
\begin{tabular}{lcccccc}
	\toprule
	\multirow{3}{*}{Method} & \multicolumn{2}{c}{VIG} & \multicolumn{3}{c}{VCG} & LCG\\
	\cmidrule(lr){2-3} \cmidrule(lr){4-6} \cmidrule(lr){7-7} 
	& Clustering & Modularity & Variable $\alpha_v$ & Clause $\alpha_v$ & Modularity & Modularity \\
	\midrule
	Training-24 & 0.53 $\pm$  0.08 & 0.61 $\pm$ 0.13 & 5.30 $\pm$ 3.79 & 5.14 $\pm$ 3.13 & 0.76 $\pm$ 0.08 & 0.70 $\pm$  0.07 \\
	G2SAT & 0.41 $\pm$ 0.18 (23\%) & {\bf 0.55 $\pm$ 0.18} (10\%) & {\bf 5.30 $\pm$ 3.79 (0\%)} & 7.22 $\pm$ 6.38 (40\%) & {\bf 0.71 $\pm$ 0.12 (7\%)} & {\bf 0.68 $\pm$ 0.06 (3\%)} \\
	\modelshort{}-0.1 & 0.49 $\pm$ 0.21 (8\%) & 0.36 $\pm$ 0.21 (41\%) & {\bf 5.30 $\pm$ 3.79 (0\%)} & 3.76 $\pm$ 1.21 (27\%) & 0.58 $\pm$ 0.16 (24\%) & 0.58 $\pm$ 0.11 (17\%) \\
	\modelshort-0.01 & {\bf 0.54 $\pm$ 0.13(2\%)} & 0.53 $\pm$ 0.21 (13\%) & {\bf 5.30 $\pm$ 3.79(0\%)} & {\bf 4.28 $\pm$ 1.50 (17\%)} & {\bf 0.71 $\pm$ 0.13 (7\%)} & 0.67 $\pm$ 0.09 (4\%) \\
	\bottomrule
\end{tabular}%
}
\vspace{-1mm}
\caption{Training and generated graph statistics with 24 SAT formulas used in \citet{you2019g2sat}. The neural baselines in \tabref{tab:benchmark} are not applicable due to scalability issue. We report mean and std of different test statistics, as well as the gap between true SAT instances.  \label{tab:g2sat_24}}
\vspace{-3mm}
\end{table*}

\begin{figure*}[t]
\noindent\begin{minipage}{0.33\textwidth}
\centering
	\includegraphics[width=1.0\textwidth]{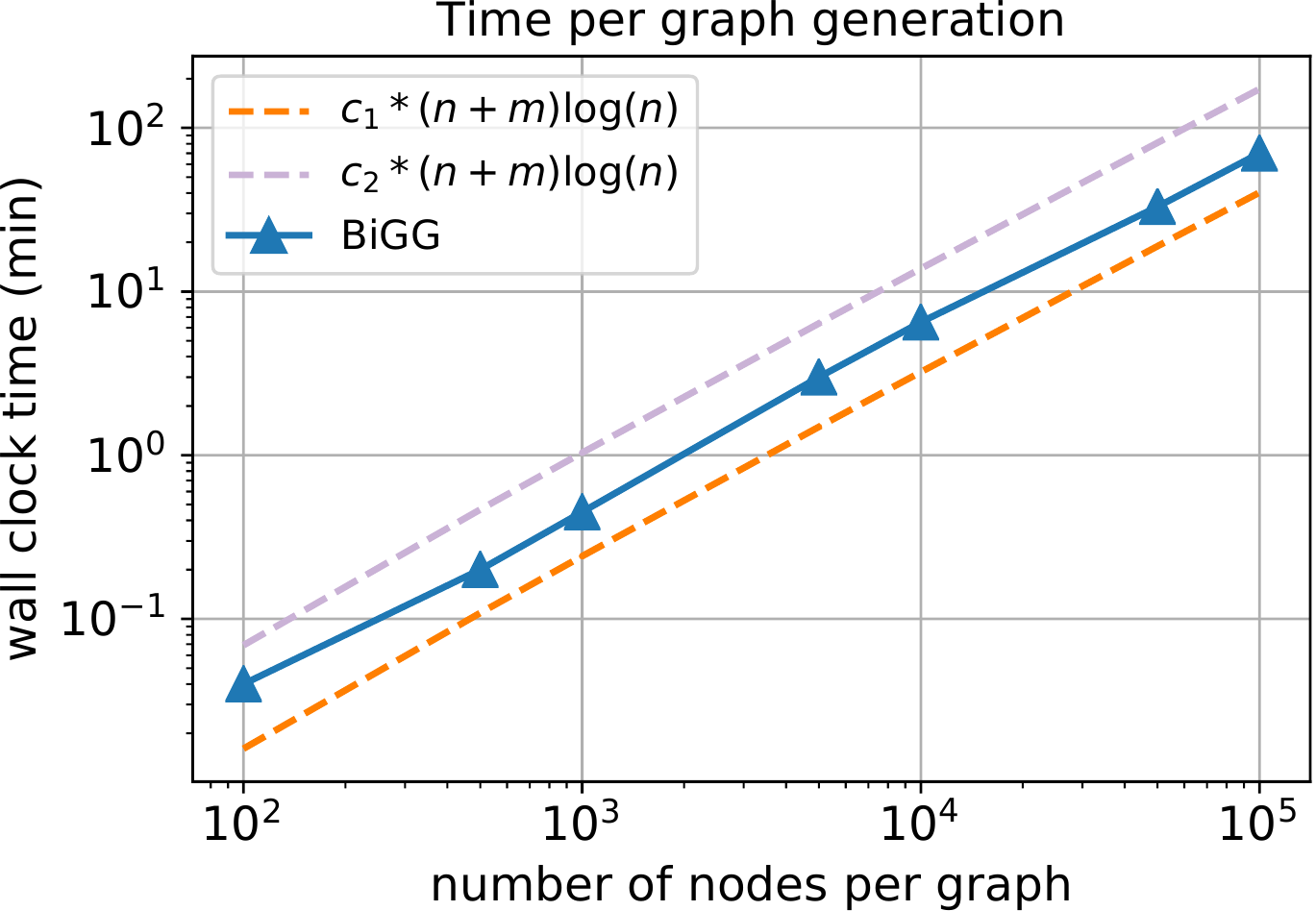}
	\vspace{-8mm}
	\caption{Inference time per graph. \label{fig:test_time}}
	\vspace{-3mm}
\end{minipage}%
\hfill
\noindent\begin{minipage}{0.33\textwidth}
\centering
	\includegraphics[width=1.0\textwidth]{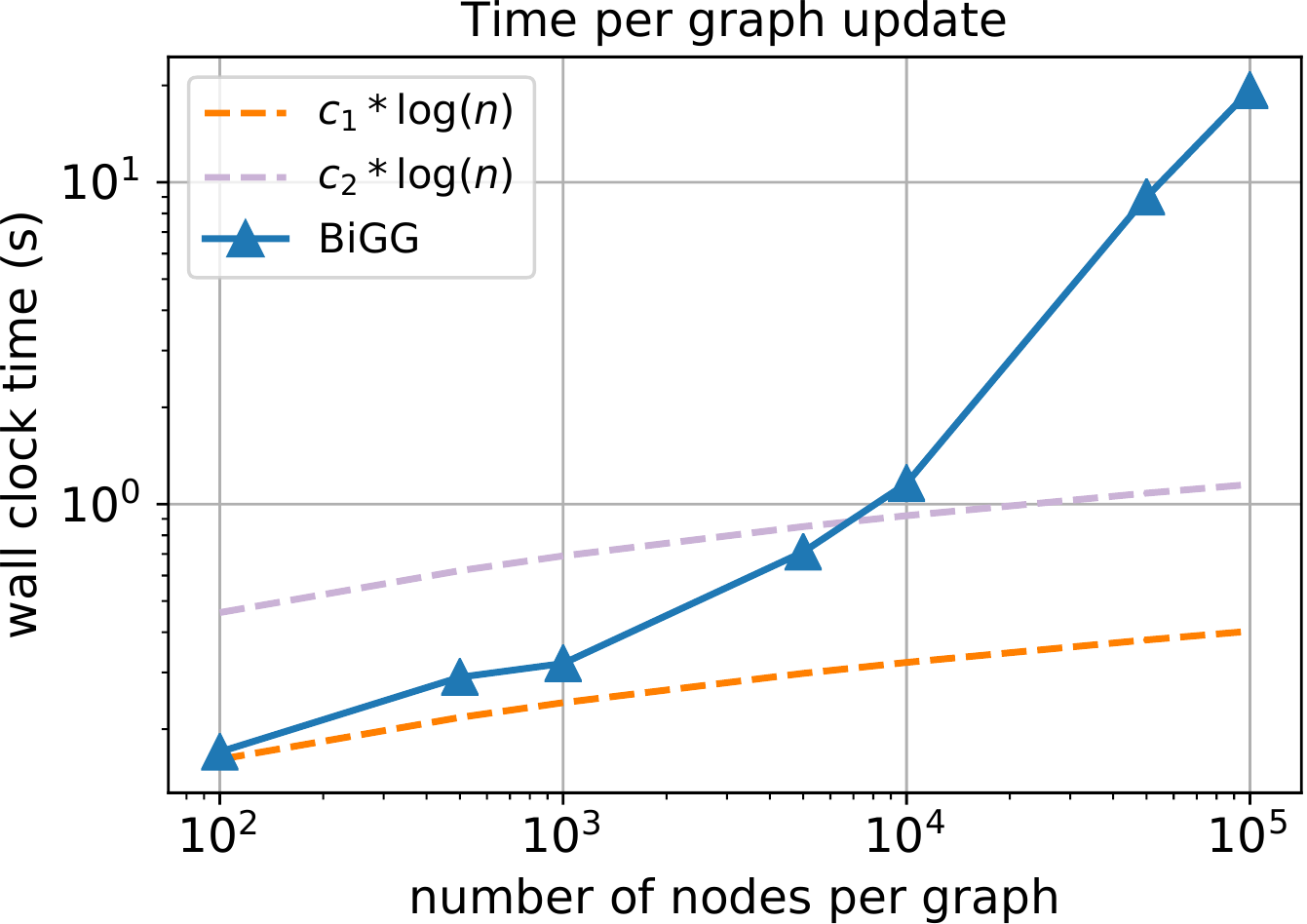}
	\vspace{-8mm}
	\caption{Training time per update. \label{fig:train_time}}
	\vspace{-3mm}
\end{minipage}%
\hfill
\noindent\begin{minipage}{0.33\textwidth}
\centering
	\includegraphics[width=1.0\textwidth]{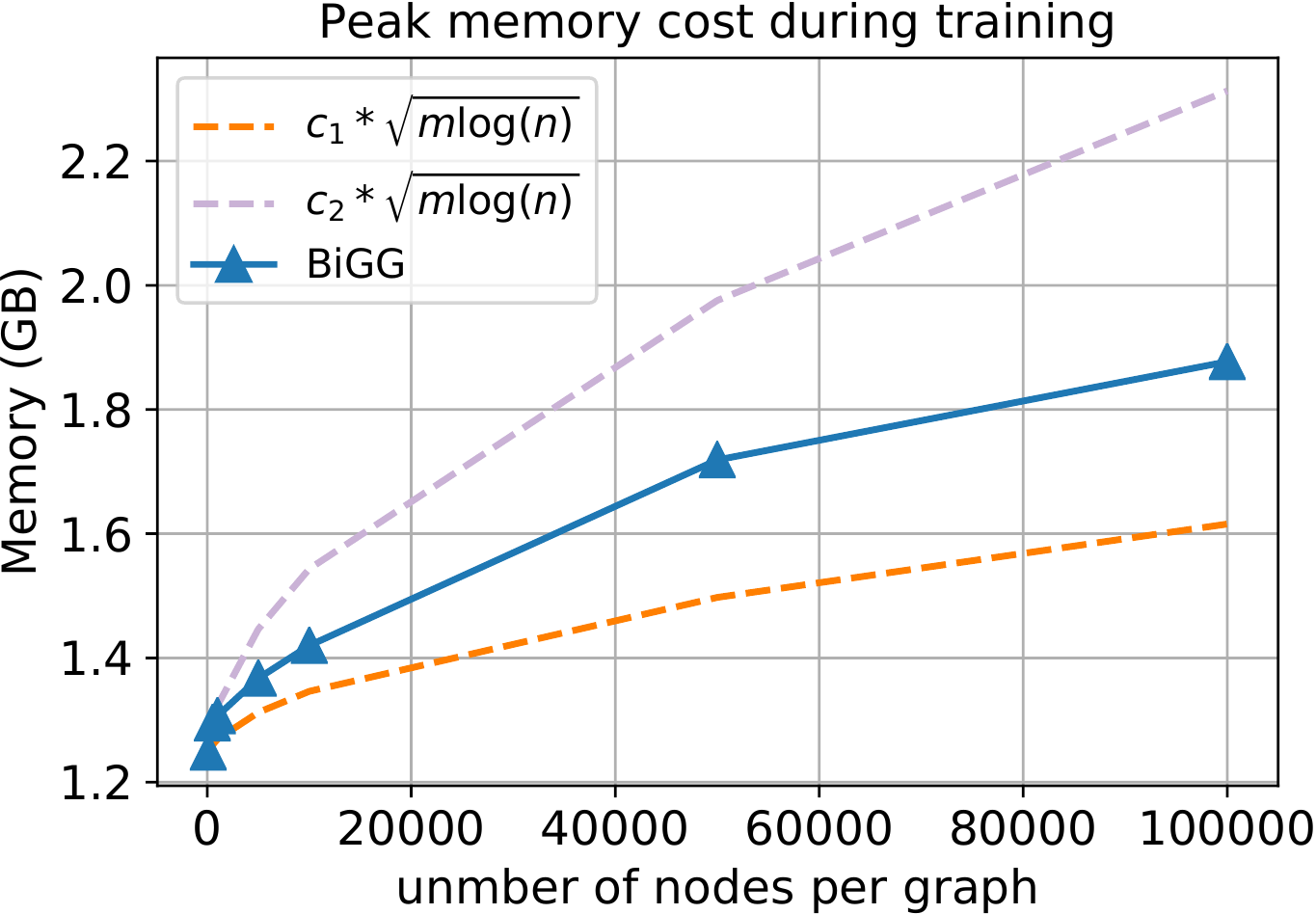}
	\vspace{-8mm}
	\caption{Training memory cost. \label{fig:memory_cost} }
	\vspace{-3mm}
\end{minipage}%
\end{figure*}

In addition to comparing with general graph generative models, in this section we compare against several models that are designated for generating the Boolean Satisfiability (SAT) instances. A SAT instance can be represented using bipartite graph, \ie, the  literal-clause graph (LCG). For a SAT instance with $n_x$ variables and $n_c$ clauses, it creates $n_x$ positive and negative literals, respectively. The canonical node ordering assigns 1 to $2*n_x$ for literals and $2*n_x+1$ to $2 * n_x + n_c$ for clauses. 

The following experiment largely follows G2SAT~\citep{you2019g2sat}. We use the train/test split of SAT instances obtained from G2SAT website. This result in 24 and 8 training/test SAT instances, respectively. The size of the SAT graphs ranges from 491 to 21869 nodes. 
Note that the original paper reports results using 10 small training instances instead. For completeness, we also include such results in~\appref{app:sat} together with other baselines from~\citet{you2019g2sat}.

\noindent\textbf{Baseline:} We mainly compare the learned model with G2SAT, a specialized deep graph generative model for bipartite SAT graphs. Since \modelshort{} is general purposed, to guarantee the generated adjacency matrix $A$ is bipartite, we let our model to generate the upper off-diagonal block of the adjacency matrix only, \ie, $A[0:2*n_x, 2*n_x:2*n_x+n_c]$.

G2SAT requires additional `template graph' as input when generating the graph. Such template graph is equivalent to specify the node degree of literals in LCG. We can also enforce the degree of each node $|\Ncal_v|$ in our model. 

\noindent\textbf{Evaluation:} Following G2SAT, we report the mean and standard deviation of statistics with respect to different test functions. These include the modularity, average clustering coefficient and the scale-free
structure parameters for different graph representations of SAT instances. Please refer to~\citet{newman2001clustering, newman2006modularity, ansotegui2009structure, clauset2009power} for more details. In general, the closer the statistical estimation the better it is. 

\noindent\textbf{Results:} Following~\citet{you2019g2sat}, we compare the statistics of graphs with the training instances in~\tabref{tab:g2sat_24}. To mimic G2SAT which picks the best action among sampled options each step, we perform $\epsilon$-sampling variant (which is denoted \modelshort-$\epsilon$). Such model has $\epsilon$ probability to sample from Bernoulli distribution (as in Eq~\eqref{eq:plch}~\eqref{eq:prch}) each step, and $1-\epsilon$ to pick best option otherwise. 
This is used to demonstrate the capacity of the model. We can see that the proposed \modelshort{} can mostly recover the statistics of training graph instances. This implies that despite being general, the full autoregressive model is capable of modeling complicated graph generative process. We additionally report the statistics of generated SAT instances against the test set in ~\appref{app:sat}, where G2SAT outperforms \modelshort{} in 4/6 metrics. As G2SAT is specially designed for bipartite graphs, the inductive bias it introduces allows the extrapolation to large graphs. Our \modelshort{} is general purposed and has higher capacity, thus also overfit to the small training set more easily.

\subsection{Scalability of \modelshort{}}
\label{sec:exp_scalability}

In this section, we will evaluate the scalability of \modelshort{} regarding the time complexity, memory consumption and the quality of generated graphs with respect to the number of nodes in graphs. 

\subsubsection{Runtime and memory cost}

Here we empirically verify the time and memory complexity analyzed in \secref{sec:model}. We run \modelshort{} on grid graphs with different numbers of nodes $n$ that are chosen from $\cbr{100, 500, 1k, 5k, 10k, 50k, 100k}$. In this case $m = \Theta(n)$. Additionally, we also plot curves from the theoretical analysis for verification. Specifically, suppose the asymptotic cost function is $f(n, m)$ w.r.t. graph size, then if there exist constants $c_1, c_2, n', m'$ such that $c_1 g(n, m) < f(n, m) < c_2 g(n, m), \forall n>n', m > m'$, then we can claim $f(n, m) = \Theta(g(n, m))$. In \figref{fig:test_time} to \ref{fig:memory_cost}, the two constants $c_1, c_2$ are tuned for better visualization. 

\figref{fig:test_time} reports the time needed to sample a single graph from the learned model. We can see the computation cost aligns well with the ideal curve of $O((n+m)\log n)$.  

To evaluate the training time cost, we report the time needed for each round of model update, which consists of forward, backward pass of neural network, together with the update of parameters. As analyzed in \secref{sec:par_train}, if there is a device with infinite FLOPS, then the time cost would be $O(\log n)$. We can see from \figref{fig:train_time} that this analysis is consistent when graph size is less than 5,000. However as graph gets larger, the computation time grows linearly on a single GPU due to the limit of FLOPS and RAM. 

Finally \figref{fig:memory_cost} shows the peak memory cost during training on a single graph. We select the optimal number of chunks $k^*=O(\sqrt{\frac{m}{\log n}})$ as suggested in \secref{sec:sqrtn}, and thus the peak memory grows as $O(\sqrt{m \log n})$. We can see such sublinear growth of memory can scale beyond sparse graphs with 100k of nodes.   

\subsubsection{Quality w.r.t graph size}

\begin{table}
\centering
\resizebox{0.48\textwidth}{!}{%
\begin{tabular}{lcccccc}
	\toprule
	 & 0.5k & 1k & 5k & 10k & 50k & 100k \\
	\midrule
	Erdős–Rényi & 0.84  & 0.86 & 0.91 & 0.93 &  0.95 & 0.95 \\
	GRAN & $2.95e^{-3}$ & $1.18e^{-2}$ & 0.39 & 1.06 & N/A & N/A \\
	\modelshort{} & $\mathbf{3.47e^{-4}}$ & $\mathbf{7.94e^{-5}}$ & $\mathbf{1.57e^{-6}}$ & $\mathbf{6.39e^{-6}}$ & $\mathbf{6.06e^{-4}}$ &  $\mathbf{2.54e^{-2}}$ \\
	\bottomrule
\end{tabular}
}
\caption{MMD using orbit test function on grid graphs with different average number of nodes. N/A denotes runtime error during training, due to RAM or file I/O limitations. \label{tab:large_grid}}
\end{table}

In addition to the time and memory cost, we are also interested in the generated graph quality as it gets larger. To do so, we follow the experiment protocols in \secref{sec:exp_general_graph} on a set of grid graph datasets. The datasets have the average number of nodes ranging in $\cbr{0.5k, 1k, 5k, 10k, 50k, 100k}$. We train on $80$ of the instances, and evaluate results on $20$ held-out instances. As calculating spectra is no longer feasible for large graphs, we report MMD with orbit test function in \tabref{tab:large_grid}. For neural generative models we compare against GRAN as it is the most scalable one currently. GRAN fails on training graphs beyond 50k nodes as runtime error occurs due to RAM or file I/O limitations. We can see the proposed \modelshort{} still preserves high quality up to grid graphs with 100k nodes. With the latest advances of GPUs, we believe \modelshort{} would scale further due to its superior asymptomatic complexity over existing methods.

\subsection{Ablation study}
\label{sec:ablation}

In this section, we take a deeper look at the performance of~\modelshort{} with different node ordering in \secref{sec:ordering}. We also show the effect of edge density to the generative performance in \secref{sec:sparsity}.

\subsubsection{\modelshort{} with different node ordering}
\label{sec:ordering}

In the previous sections we use DFS or BFS orders. We find these two orders give consistently good performance over a variety of datasets. For the completeness, we also present results with other node orderings. 

We use different orders presented in GRAN's \texttt{Github} implementation. We use the protein dataset with spectral-MMD as evaluation metric. See~\tabref{tab:ordering} for the experimental results. In summary: 1) BFS/DFS give our model consistently good performance over all tasks, as it reduces the tree-width for BiGG (similar to Fig5 in GraphRNN) and we suggest to use BFS or DFS by default; 2) BiGG is also flexible enough to take any order, which allows for future research on deciding the best ordering.

\begin{table}[h]
\centering
\resizebox{0.48\textwidth}{!}{%
\begin{tabular}{cccccc}
	\toprule
	DFS & BFS & Default & Kcore & Acc & Desc \\
	\hline
	3.64$e^{-3}$ & 3.89$e^{-3}$ & 4.81$e^{-3}$ & 2.60$e^{-2}$ & 3.93$e^{-3}$ & 4.54$e^{-3}$ \\
	\bottomrule
\end{tabular}
}
\caption{\modelshort{} with nodes ordered by DFS, BFS, default, k-core ordering, degree accent ordering and degree descent ordering respectively on protein data. We report spectral-MMD metric here. \label{tab:ordering}}
\end{table}
Determining the optimal ordering is NP-hard, and learning a good ordering is also difficult, as shown in the prior works. In this paper, we choose a single canonical ordering among graphs, as \citet{li2018learning} shows that canonical ordering mostly outperforms variable orders in their Table 2, 3, while \citet{liao2019efficient} uses single DFS ordering (see their Sec 4.4 or github) for all experiments. 

\subsubsection{Performance on random graphs with decreasing sparsity}
\label{sec:sparsity}

 We here present experiments on Erdos-Renyi graphs with on average 500 nodes and different densities. We report spectral MMD metrics for GRAN, GT and BiGG, where GT is the ground truth Erdos-Renyi model for the data. 

\begin{table}[h]
\centering
\begin{tabular}{ccccc}
	\toprule
	 & 1\% & 2\% & 5\% & 10\%  \\
	\hline
	GRAN & 3.50$e^{-1}$ & 1.23$e^{-1}$ & 7.81$e^{-2}$ & 1.31$e^{-2}$\\
	GT & 9.97$e^{-4}$ & 4.55 $e^{-4}$ & 2.82$e^{-4}$ & 1.94$e^{-4}$ \\
	\modelshort & 9.47$e^{-4}$ & 5.08$e^{-4}$ & 3.18$e^{-4}$ & 8.38$e^{-4}$\\
	\bottomrule 
\end{tabular}
\caption{Graph generation quality with decreasing sparsity. We use spectral-MMD as evaluation metric against held-out test graphs. \label{tab:sparsity}}
\end{table}

Our main focus is on sparse graphs that are more common in the real world, and for which our approach can gain significant speed ups over the alternatives. Nevertheless, as shown in~\tabref{tab:sparsity}, we can see that BiGG is consistently doing much better than GRAN while being close to the ground truth across different edge densities.

\section{Conclusion}
\label{sec:conclusion}

We presented \modelshort{}, a scalable autoregressive generative model for general graphs. It takes $O((n+m) \log n)$ complexity for sparse graphs, which substantially improves previous $\Omega(n^2)$ algorithms. We also proposed both time and memory efficient parallel training method that enables comparable or better quality on benchmark and large random graphs. Future work include scaling up it further while also modeling attributed graphs. 

\clearpage

\section*{Acknowledgements}

We would like to thank Azalia Mirhoseini, Polo Chau, Sherry Yang and anonymous reviewers for valuable comments and suggestions.

\appendix
\onecolumn

\section{Detailed discussion of related work}
\label{app:related}

\paragraph{Traditional graph generative models }
There has been a lot of work~\citep{Erdos1959, barabasi1999emergence, Barabasi2002complex, chakrabarti2004r, robins2007exprandom, Leskovec2010kronecker, Airoldi2008mixed} on generating graphs with a set of specific properties like degree distribution, diameter, and eigenvalues.
While some of them~\citep{Erdos1959, barabasi1999emergence, Barabasi2002complex, chakrabarti2004r} focused on the statistical mechanics of random and complex networks, \citep{robins2007exprandom} proposed exponential random graph models, known as $p\ast$ model, in estimating network models comes from the area of social sciences, statistics and social network analysis. However the $p\ast$ model usually focuses on local structural features of networks. \citep{Airoldi2008mixed, Leskovec2010kronecker}, on the other hand, model the structure of the network as a whole by combining a global model of dense patches of connectivity with a local model. 

Most of these classical models are hand-engineered to model a particular family of graphs, and thus do not have the capacity to directly learn the generative model from observed data.

\paragraph{Deep graph generative models }
Recent work on deep neural network based graph generative models are typically much more expressive than classic random graph models, at a higher computation cost.

Similar to visual data, there are VAE \cite{kingma2013auto} and GAN \cite{goodfellow2014generative} based graph generative models, like GraphVAEs \cite{kipf2016variational,simonovsky2018graphvae} and NetGAN \cite{bojchevski2018netgan}.  These models typically generates large parts (or all) of the adjacency matrix independently, making them unable to model complex graph structures.

Auto-regressive models, on the other hand, generate a graph sequentially part by part, gaining expressivity but are typically more expensive as large graphs require large numbers of generation steps.
Scalability is a constant important topic in this line of work, starting from \cite{li2018learning}.  More recent work GraphRNN \cite{you2018graphrnn} and GRAN \cite{liao2019efficient} reported improved scalability and success generating graphs of up to 5,000 nodes.  Our novel \modelshort{} model advances this line of work by significantly improving both on model quality and scalability.

Besides models for general graphs, a lot of work also exploit domain knowledge for better performance in specific domains.  Examples of this include \citep{kusner2017grammar,dai2018syntax,jin2018junction,liu2018constrained} for modeling molecule graphs, and \cite{you2019g2sat} for SAT instances.

\section{More experiment details}
\label{app:exp}

\paragraph{Manual Batching}

Though theoretically \secref{sec:par_train} enables the parallelism, the commonly used packages like PyTorch and TensorFlow has limited support for auto-batching operators. Also optimizing over a computation graph with $O(m)$ operators directly would be infeasible. 
Inspired by previous works~\citep{looks2017deep, neubig2017fly, bradbury2018automatic}, we enable the parallelism by performing gathering and dispatching for tree or LSTM cells. This requires a single call of \texttt{gemm} on CUDA, instead of multiple \texttt{gemv} calls for the cell operations. The indices used for gathering and dispatching are computed in a customized c++ op for the sake of speed.

\paragraph{Parameter sharing}

As the maximum depth in~\modelshort{} is $O(\log n)$, it is feasible to have different parameters per different layers for tree or LSTM cells. However experimentally it didn’t affect the performance significantly. With this change, not all parameters are updated at each step, as the trees have different heights and some are updated more often than others, which could potentially make the learning harder. Also it would make the model $O(\log n)$ times larger, and limits the potential of generating graphs larger than seen during training. So we still share the parameters of cells like other autoregressive models.

\subsection{\modelshort{} setup}

\noindent\textbf{Hyperparameters:} For the proposed~\modelshort{}, we use embedding size $d=256$ together with position encoding for state representation. Learning rate is initialized to $1e^{-3}$ and decays to $1e^{-5}$ when training loss gets plateau. 

All the experiments for both baselines and \modelshort{} are carried out on a single V100 GPU. For the sublinear memory technique mentioned in~\secref{sec:sqrtn}, we choose $k$ (\ie, the number of blocks) as small as possible such that the training would be able to fit in a single V100 GPU. This is to make the training feasible while minimizing the synchronization overhead. Empirically the block size used in sublinear memory technique is set to 6,000 for sparse graphs. 

\subsection{Baselines setup}

We include all the baseline numbers from their original papers when applicable, as the most experimental protocols are the same. We run the following baselines when the corresponding numbers are not reported in the literature:
\begin{itemize}
	\item Erdős–Rényi: for the results in \tabref{tab:large_grid}, we estimate the graph edge density using the training grid graphs, and generate new Erdős–Rényi random graphs with this density and compare with the held-out test graphs; 
	\item GraphRNN-S~\citep{you2018graphrnn}: the result on Lobster random graphs in~\tabref{tab:benchmark} is obtained by training the GraphRNN using official code~\footnote{\url{https://github.com/JiaxuanYou/graph-generation}} for 3000 epochs; 
	\item GRAN~\citep{liao2019efficient}: the results of GRAN in \tabref{tab:large_grid} are obtained by training GRAN with grid graph configuration for 8,000 epochs or 3 days, whichever limit the model reaches first. 
	\item G2SAT~\citep{you2019g2sat}: For results in~\tabref{tab:g2sat_24} and ~\tabref{tab:g2sat_24_app}, we run the code~\footnote{\url{https://github.com/JiaxuanYou/G2SAT}} for 1000 epochs. For the test graph comparison in~\tabref{tab:g2sat_24_app}, we use the 8 test graphs as `template-graphs' for G2SAT to generate graphs.
\end{itemize}

\begin{table*}[t]
\centering
\resizebox{1.0\textwidth}{!}{%
\begin{tabular}{lcccccc}
	\toprule
	\multirow{3}{*}{Method} & \multicolumn{2}{c}{VIG} & \multicolumn{3}{c}{VIG} & LCG\\
	\cmidrule(lr){2-3} \cmidrule(lr){4-6} \cmidrule(lr){7-7} 
	& Clustering & Modularity & Variable $\alpha_v$ & Clause $\alpha_v$ & Modularity & Modularity \\
	\midrule
	Training & 0.50 $\pm$ 0.07 & 0.58 $\pm$ 0.09 & 3.57 $\pm$ 1.08 & 4.53 $\pm$ 1.09 & 0.74 $\pm$ 0.06 & 0.67 $\pm$ 0.05 \\
	CA & 0.33 $\pm$ 0.08 (34\%) & 0.48 $\pm$ 0.10 (17\%) & 6.30 $\pm$ 1.53 (76\%) &  N/A  & 0.65 $\pm$ 0.08 (12\%) & 0.53 $\pm$ 0.05 (16\%) \\
	PS(T=0) & 0.82 $\pm$ 0.04 (64\%) & 0.72 $\pm$ 0.13 (24\%) & 3.25 $\pm$ 0.89 (9\%) & 4.70 $\pm$ 1.59 (4\%) & 0.86 $\pm$ 0.05 (16\%) & {\bf 0.64 $\pm$ 0.05 (2\%)} \\
	PS(T=1.5) & 0.30 $\pm$ 0.10 (40\%) & 0.14 $\pm$ 0.03 (76\%) & 4.19 $\pm$ 1.10 (17\%) & 6.86 $\pm$ 1.65 (51\%) & 0.40 $\pm$ 0.05 (46\%) & 0.41 $\pm$ 0.05 (35\%) \\
	G2SAT & 0.41 $\pm$ 0.09 (18\%) & 0.54 $\pm$ 0.11 (7\%) & {\bf 3.57 $\pm$ 1.08 (0\%)} & 4.79 $\pm$ 2.80 (6\%) & 0.68 $\pm$ 0.07 (8\%) & 0.67 $\pm$ 0.03 (6\%) \\
	\modelshort\ & {\bf 0.50 $\pm$ 0.07 (0\%)} & {0.58 $\pm$ 0.09 (0\%)} & {\bf 3.57 $\pm$ 1.08 (0\%)} & {\bf 4.53 $\pm$ 1.09 (0\%)} & 0.73 $\pm$ 0.06 (1\%) & 0.61 $\pm$ 0.09 (9\%) \\
	\modelshort{}-0.1 & {\bf 0.50 $\pm$ 0.07 (0\%)} & {\bf 0.58 $\pm$ 0.09 (0\%)} & {\bf 3.57 $\pm$ 1.08 (0\%)} & {\bf 4.53 $\pm$ 1.09 (0\%)} & {\bf 0.74 $\pm$ 0.06 (0\%)} & 0.65 $\pm$ 0.07 (7\%) \\
	\bottomrule
\end{tabular}%
}
\caption{Training and generated graph statistics with 10 SAT formulas used in \citet{you2019g2sat}. Baselines results are directly copied from \citet{you2019g2sat}. \label{tab:subset}}
\end{table*}

\begin{table*}[t]
\centering
\resizebox{1.0\textwidth}{!}{%
\begin{tabular}{lcccccc}
	\toprule
	\multirow{3}{*}{Method} & \multicolumn{2}{c}{VIG} & \multicolumn{3}{c}{VIG} & LCG\\
	\cmidrule(lr){2-3} \cmidrule(lr){4-6} \cmidrule(lr){7-7} 
	& Clustering & Modularity & Variable $\alpha_v$ & Clause $\alpha_v$ & Modularity & Modularity \\
	\midrule
	Test-8 & 0.50 $\pm$ 0.05 & 0.68 $\pm$ 0.19 & 4.66 $\pm$ 1.92 & 6.33 $\pm$ 3.23 & 0.84 $\pm$ 0.02 & 0.75 $\pm$ 0.05\\
	G2SAT & 0.22 $\pm$ 0.10 (56\%)  & {\bf 0.74 $\pm$ 0.11 (9\%)} & {\bf 4.66 $\pm$ 1.92 (0\%)}  & {\bf 6.60 $\pm$ 4.15 (5\%) } & {\bf 0.80 $\pm$ 0.09 (5\%)} & {\bf 0.68 $\pm$ 0.04 (9\%)} \\
	\modelshort{} & {\bf 0.37 $\pm$ 0.19 (26\%)} & 0.28 $\pm$ 0.12 (58\%) & {\bf 4.66 $\pm$ 1.92 (0\%)} & 2.74 $\pm$ 0.37 (57\%) & 0.44 $\pm$ 0.07 (48\%) & 0.48 $\pm$ 0.09 (36\%) \\
	\bottomrule
\end{tabular}%
}
\caption{Test and generated graph statistics with 8 SAT formulas from~\url{https://github.com/JiaxuanYou/G2SAT}. G2SAT and \modelshort{} are trained on 24 sat formulas as in~\tabref{tab:g2sat_24} \label{tab:g2sat_24_app}} 
\end{table*}

\subsection{More experimental results on SAT graphs}
\label{app:sat}
In main paper we reported the results on 24 training instances with the evaluation metric used in~\citet{you2019g2sat}. Here we include the results on the subset which is used in the original G2SAT paper~\citep{you2019g2sat}. This subset contains 10 instances, with 82 to 1122 variables and 327 to 4555 clauses. This result in graphs with 6,799 nodes at most. \tabref{tab:subset} summarizes the results, together with other baseline results that are copied from~\citet{you2019g2sat}. We can see our model can almost perfectly recover the training statistics if the generative process is biased more towards high probability region (\ie, \modelshort{}-0.1 which has 90\% chance to use greedy decoding at each step). This again demonstrate that our model is flexible enough to capture complicated distributions of graph structures. On the other hand, as~\tabref{tab:g2sat_24_app} shows, the G2SAT beats \modelshort{} on 4 out of 6 metrics. Since G2SAT is a specialized generative model for bipartite graphs, it enjoys more of inductive bias towards this specific problem, and thus would work better in the low-data and extrapolation scenario. Our general purposed model thus would be expected to require more training data due to its high capacity in model space. 

\end{document}